\documentclass{article}
\usepackage{ijcai16}

\usepackage{times}

\usepackage{helvet}
\usepackage{courier}
\usepackage{url}
\usepackage{graphicx} 
\usepackage{subfigure} 
\usepackage{algorithm}
\usepackage{algorithmic}
\usepackage{amsmath}[11pt]
\usepackage{amsmath, amsthm, amssymb}
\usepackage{amsthm}
\usepackage{amsfonts}[11pt]
\usepackage{mathtools}
\usepackage{bbm}

\newcommand{\bc}{\begin{center}}
\newcommand{\ec}{\end{center}}
\newcommand{\bq}{\begin{quote}}
\newcommand{\eq}{\end{quote}}

\newcommand{\be}{\begin{equation}}
\newcommand{\ee}{\end{equation}}
\newcommand{\beqa}{\begin{eqnarray*}}
\newcommand{\eeqa}{\end{eqnarray*}}
\newcommand{\beqn}{\begin{eqnarray}}
\newcommand{\eeqn}{\end{eqnarray}}
\newcommand{\bbibl}{\begin{thebibliography}{99}}
\newcommand{\ebibl}{\end{thebibliography}}

\newcommand{\ba}{\begin{array}}
\newcommand{\ea}{\end{array}}

\DeclareMathOperator*{\argmax}{argmax}

\newcommand{\E}{\mathbb{E}}

\begin{document}
%
\title{Personalized Advertisement Recommendation: A Ranking Approach to Address the Ubiquitous Click Sparsity Problem}
\author{
Sougata Chaudhuri \\
Department of Statistics\\
University of Michigan, Ann Arbor\\
\And
Georgios Theocharous \\
Adobe Big Data Experience Lab\\
\And
Mohammad Ghavamzadeh \\
Adobe Big Data Experience Lab\\
}

\maketitle


\begin{abstract}
We study the problem of personalized advertisement recommendation (PAR), which consist of a user visiting a system (website) and the system displaying one of $K$ ads to the user. The system uses an internal ad recommendation policy to map the user's profile (context) to one of the ads. The user either clicks or ignores the ad and correspondingly, the system updates its recommendation policy. PAR problem is usually tackled by scalable \emph{contextual bandit} algorithms, where the policies are generally based on classifiers. A  practical problem in PAR is extreme click sparsity, due to very few users actually clicking on ads. We systematically study the drawback of using contextual bandit algorithms based on classifier-based policies, in face of extreme click sparsity. We then suggest an alternate policy, based on rankers, learnt by optimizing the Area Under the Curve (AUC) ranking loss, which can significantly alleviate the problem of click sparsity. We conduct extensive experiments on public datasets, as well as three industry proprietary datasets, to illustrate the improvement in click-through-rate (CTR) obtained by using the ranker-based policy over classifier-based policies.
\end{abstract}


\section{Introduction}
Personalized advertisement recommendation (PAR) system is intrinsic to many major tech companies like Google, Yahoo, Facebook and others. The particular PAR setting we study here consists of a policy that displays one of the $K$ possible ads/offers, when a user visits the system. The user's profile is represented as a context vector, consisting of relevant information like demographics, geo-location, frequency of visits, etc. Depending on whether user clicks on the ad, the system gets a reward of value $1$, which in practice translates to dollar revenue. The policy is (continuously) updated from historical data, which consist of tuples of the form $\{\text{{\em user context}, {\em displayed ad}, {\em reward}}\}$. We will, in this paper, concern ourselves with PAR systems that are geared towards maximizing total number of clicks. 

The plethora of papers written on the PAR problem makes it impossible to provide an exhaustive list. Interested readers may refer to a recent paper by a team of researchers in Google~\cite{mcmahan2013ad} and references therein. While the techniques in different papers differ in their details, the majority of them can be be analyzed under the umbrella framework of \emph{contextual bandits}~\cite{langford2008epoch}. The term \emph{bandit} refers to the fact that the system only gets to see the user's feedback on the ad that was displayed, and not on any other ad.  
Bandit information leads to difficulty in estimating the expected reward of a new or a relatively unexplored ad (the \emph{cold start} problem). Thus, contextual bandit algorithms, during prediction, usually balance between \emph{exploitation} and \emph{exploration}. Exploitation consists of predicting according to the current recommendation policy, which usually selects the ad with the maximum estimated reward, and exploration consists of systematically choosing some other ad to display, to gather more information about it.

Most contextual bandit algorithms aim to learn a policy that is essentially some form of multi-class classifier. For example, one important class of contextual bandit algorithms learn a classifier per ad from the batch of data \cite{richardson2007predicting,mcmahan2013ad,he2014practical} and convert it into a policy, that displays the ad with the highest classifier score to the user (exploitation). Some exploration techniques, like explicit $\epsilon$-greedy~\cite{koh2014empirical,theocharous2015ad} or implicit Bayesian type sampling from the posterior distribution maintained on classifier parameters~\cite{chapelle2011empirical} are sometimes combined with this exploitation strategy. Other, more theoretically sophisticated online bandit algorithms, essentially learn a cost-sensitive multi-class classifier by updating after every round of user-system interaction~\cite{dudik2011efficient,agarwal2014taming}.

Despite the fact that PAR has always been mentioned as one of the main applications of CB algorithms, there has not been much investigation into the practical issues raised in using classifier-based policies for PAR. The potential difficulty in using such policies in PAR stems from the problem of \emph{click sparsity}, i.e.,~very few users actually ever click on online ads and this lack of positive feedback makes it difficult to learn good classifiers. Our main objective here is to study this important practical issue and we list our contributions:
\begin{itemize}
\item We detail the framework of contextual bandit algorithms and discuss the problem associated with click sparsity.
\item We suggest a simple ranker-based policy to overcome the click sparsity problem. The rankers are learnt by optimizing the \emph{Area Under Curve} (AUC) ranking loss via \emph{stochastic gradient descent} (SGD)~\cite{shamir2013stochastic}, leading to a highly scalable algorithm. The rankers are then combined to create a recommendation policy.
\item We conduct extensive experiments to illustrate the improvement provided by our suggested method over both linear and ensemble classifier-based policies for the PAR problem. Our first set of experiments compare \emph{deterministic policies} on publicly available classification datasets, that are converted to bandit datasets following standard techniques. Our second set of experiments compare \emph{stochastic policies} on three proprietary bandit datasets, for which we employ a high confidence offline contextual bandit evaluation technique.
\end{itemize}


\section{Contextual Bandit (CB) Approach to PAR}
The main contextual bandit algorithms can be largely divided into two classes: those that make specific parametric assumption about the reward generation process and those that simply assume that context and rewards are generated i.i.d.~from some distribution. The two major algorithms in the first domain are LinUCB~\cite{chu2011contextual} and Thompson sampling~\cite{agrawal2013thompson}. Both algorithms assume that the reward of each ad (arm) is a continuous linear function of some unknown parameter, which is not a suitable assumption for click-based \emph{binary} reward in PAR. Moreover, both algorithms assume that there is context information available for each ad, while we assume availability of only user context in our setting. Thus, from now on, we focus on the second class of the contextual bandit algorithms. We provide a formal description of the framework of contextual bandits suited to the PAR setting, and then discuss the problem that arises due to click sparsity.

Let $\mathcal{X} \subseteq \mathbb{R}^d$ and $[K]=\{1,2,\ldots,K\}$ denote the user context space and $K$ different ads/arms. At each round, it is assumed that a pair $(x,r)$ is drawn i.i.d. from some unknown joint distribution $D$ over $\mathcal{X} \times \{0,1\}^K$. Here, $x \in \mathcal{X}$ and $r \in \{0,1\}^K$ represent the user context vector and the full reward vector, i.e.,~the user's true preference for all the ads (the full reward vector is unknown to the algorithm). $\Pi$ is the space of policies such that for any $\pi \in \Pi$, $\pi:\mathcal{X} \mapsto [K]$. Contextual bandit algorithms have the following steps:
\begin{itemize}
\item At each round $t$, the context vector $x_t$ is revealed, i.e.,~a user visits the system.
\item The system selects ad $a_t$ according to the current policy $\pi_t \in \Pi$ (exploitation strategy). Optionally, an exploration strategy is sometimes added, creating a distribution $p_t(\cdot)$ over the ads and $a_t$ is drawn from $p_t(\cdot)$. Policy $\pi_t$ and distribution $p_t$ are sometimes used synonymously by considering $\pi_t$ to be a stochastic policy.
\item Reward $r_{t,a_t}$ is revealed and the new policy $\pi_{t+1} \in \Pi$ is computed, using information $\{x_t,a_t,r_{t,a_t}\}$. We \emph{emphasize} that the system does not get to know $r_{t,a'}$, $\forall \ a' \neq a_t$.
\end{itemize}
Assuming the user-system interaction happens over $T$ rounds, the objective of the system is to maximize its cumulative reward, i.e., $\sum_{t=1}^T r_{t,a_t}$. Note that since rewards are assumed to be binary, $\sum_{t=1}^T r_{t,a_t}$ is precisely the total number of clicks and $\frac{\sum_{t=1}^T r_{t,a_t}}{T}$ is the overall CTR of the recommendation algorithm. Theoretically, performance of a bandit algorithm is analyzed via the concept of \emph{regret}, i.e.,
\begin{equation*}
\text{Regret(T)}= \underbrace{\sum_{t=1}^T r_{t,\pi^*(x_t)}}_{\text{optimal policy's cumulative reward}} - \underbrace{\sum_{t=1}^T r_{t,a_t}}_{\text{algorithm's cumulative reward}}
\end{equation*} 
where $\pi^*= \underset{\pi \in \Pi}{\argmax} \ \sum_{t=1}^T r_{t,\pi(x_t)}$. The desired property of any contextual bandit algorithm is to have a sublinear (in $T$ ) bound on $\text{Regret(T)}$ (in expectation or high probability), i.e.,~$\text{Regret(T)} \le \text{o(T)}$. \emph{This guarantees that, at least, the algorithm converges to the optimal policy $\pi^*$ asymptotically}. 


\section{Practical Issues with CB Policy Space}
\label{policy-space}
Policy space $\Pi$ considered for major contextual bandit algorithms are based on classifiers. They can be tuples of binary classifiers, with one classifier per ad, or global cost-sensitive multi-class classifier, depending on the nature of the bandit algorithm. Since clicks on the ads are rare and small improvement in click-through rate can lead to significant reward, it is vital for the policy space to have good policies that can identify the correct ads for the rare users who are highly likely to click on them. Extreme click sparsity makes it \emph{very practically challenging} to design a classifier-based policy space, where policies can identify the correct ads for rare users. \emph{Crucially, contextual bandit algorithms are only concerned with  converging as fast as possible to the best policy $\pi^*$ in the policy space $\Pi$ and do not take into account the nature of the policies. Hence, if the optimal policy in the policy space does a poor job in identifying correct ads, then the bandit algorithm will have very low cumulative reward, regardless of its sophistication}.  We discuss how click sparsity hinders in the design of different types of classifier-based policies. 

\subsection{Binary Classifier Based Policies} Contextual bandit algorithms are traditionally presented as online algorithms, with continuous update of policies. Usually, in industrial PAR systems, it is highly impractical to update policies continuously, due to thousands of users visiting a system in a small time frame. Thus, policy update happens, i.e. new policy is learnt, after intervals of time, using the bandit data produced from the interaction between the current policy and users, collected in batch. It is convenient to learn a binary classifier per ad in such a setting. To explain the process concisely, we note that the bandit data consists of tuples of the form $\{x,a,r_a\}$. For each ad $a$, the users who had not clicked on the ad  ($r_a$=0) would be considered as negative examples and the users who had clicked on the ad ($r_a$=1) would be considered as positive examples, creating a binary training set for ad $a$. The $K$ binary classifiers are converted into a recommendation policy using a ``one-vs-all" method~\cite{rifkin2004defense}. \emph{Thus, each policy in policy space $\Pi$ can be considered to be a tuple of $K$ binary classifiers}.\\
A number of research publications show that researchers consider binary linear classifiers, that are learnt by optimizing the logistic loss~\cite{richardson2007predicting}, while ensemble classifiers, like random forests, are also becoming popular~\cite{koh2014empirical}. We note that the majority of the papers that learn a logistic linear classifier focus on feature selection~\cite{he2014practical}, novel regularizations to tackle high-dimensional context vectors~\cite{mcmahan2013ad}, or propose clever combinations of logistic classifiers~\cite{agarwal2009translating}. \\
Click sparsity poses difficulty in design of accurate binary classifiers in the following way: for an ad $a$, there will be very few clicks on the ad  as compared to the number of users who did not click on the ad. A binary classifier learnt in such setting will almost always predict that its corresponding ad will not be clicked by a user, failing to identify the rare, but very important, users who are likely to click on the ad. This is  colloquially referred to as ``class imbalance problem" in binary classification~\cite{japkowicz2002class}. Due to the extreme nature of the imbalance problem, tricks like under-sampling of negative examples or oversampling of positive examples~\cite{chawla2004editorial} are not  very useful. More sophisticated techniques like cost-sensitive svms require prior knowledge about importance of each class, which is not generally available in the PAR setting. \\
{\bf Note}- Some of the referenced papers do not have explicit mention of CBs because the focus in those papers is on the issues related to classifier learning process, involving type of regularization, overcoming curse of dimensionality, scalability etc. The important issue of extreme class imbalance has not received sufficient attention (Sec 6.2, \cite{he2014practical}). When the classifiers are used to predict ads, the technique is a particular CB algorithm (the exact exploration+ exploitation mix is often not revealed). 

\subsection{Cost Sensitive Multi-Class Classifier Based Policies}  Another type of policy space consist of cost-sensitive multi-class classifiers~\cite{langford2008epoch,dudik2011efficient,agarwal2014taming}. They can be cost-sensitive multi-class svms~\cite{cao2013optimized}, multi-class logistic classifiers or filter trees~\cite{beygelzimer2007multiclass}. Click sparsity poses slightly different kind of problem in practically designing a policy space of such classifiers. \\
Cost sensitive multi-class classifier works as follows: assume a context-reward vector pair (x,r) is generated as described in the PAR setting. The classifier will try to select a class (ad) $a$ such that the reward $r_a$ is maximum among all choices of $r_{a'}$, $\forall \ a' \in [K]$ (we consider reward maximizing classifiers, instead of cost minimizing classifiers). Unlike in traditional multi-class classification, where one entry of $r$ is $1$ and all other entries are $0$; in cost sensitive classification, $r$ can have any combination of $1$ and $0$. Now consider the reward vectors $r_t$s generated over $T$ rounds. A poor quality classifier $\pi^p$, which fails to identify the correct ad for most users $x_t$, will have very low average reward, i.e.,$\frac{\sum_{t=1}^T r_{t,\pi^p(x_t)}}{T} \sim O(\epsilon)$, with $\epsilon \sim$ 0. From the model perspective, extreme click sparsity translates to almost all reward vectors $r_t$ being $\vec{\bf 0}$. Thus, even a very good classifier $\pi^g$, which can identify the correct ad $a_t$ for almost all users, will have very low average reward, i.e.,~$\frac{\sum_{t=1}^T r_{t,\pi^g(x_t)}}{T} \sim O(\epsilon)$. From a practical perspective, it is difficult to distinguish between the performance of a good and poor classifier, in face of extreme sparsity, and thus, cost sensitive multi-class classifiers are not ideal policies for contextual bandits addressing the PAR problem.



\section{AUC Optimized Ranker}

We propose a ranking-based alternative to learning a classifier per ad, in the offline setting, that is capable of overcoming the click sparsity problem. We learn a ranker per ad by optimizing the Area Under the Curve (AUC) loss, and use a ranking score normalization technique to create a policy mapping context to ad. We note that AUC is a popular measure used to evaluate a classifier on an imbalanced dataset. However, our objective is to explicitly use the loss to learn a ranker that overcomes the imbalance problem and then create a context to ad mapping policy.\\
{\bf Ranker Learning Technique}: For an ad $a$, let $S^+= \{x: r_a=1\}$ and $S^-= \{x:r_a=0\}$ be the set of positive and negative instances, respectively. Let $f_w$ be a linear ranking function parameterized by $w \in \mathbb{R}^d$, i.e., $f_w(x)= w\cdot x$ (inner product). AUC-based loss (AUCL) is a ranking loss that is minimized when positive instances get higher scores than negative instances, i.e.,~the positive instances are ranked higher than the negatives when instances are sorted in descending order of their scores~\cite{cortes2004auc}. Formally, we define empirical AUCL for function $f_w(\cdot)$ 
\begin{equation*}
\small
\text{AUCL}= \dfrac{1}{|S^+||S^-|} \sum_{x^+ \in S^+} \sum_{x^- \in S^-} \mathbbm{1}(\underbrace{f_w( x^+) -  f_w(x^-)}_{t} <0).
\end{equation*}
Direct optimization of AUCL is a NP-hard problem, since AUCL is sum of discontinuous indicator functions. To make the objective function computationally tractable, the indicator functions are replaced by a continuous, convex surrogate $\ell(t)$. Examples include hinge $\ell(t)= [1-t]_{+}$ and logistic $\ell(t)= \log(1+ exp(-t))$ surrogates. Thus, the final objective function to optimize is
\begin{equation}
\small
\label{eq:objectivefn}
L(w)=  \dfrac{1}{|S^+||S^-|} \sum_{x^+ \in S^+} \sum_{x^- \in S^-} \ell (\underbrace{f_w( x^+) -  f_w( x^-)}_{t}).
\end{equation} \\
{\bf Note}: Since AUCL is a ranking loss, the concept of class imbalance ceases to be a problem. Irrespective of the number of positive and negative instances in the training set, the position of a positive instance w.r.t to a negative instance in the final ranked list is the only matter of concern in AUCL calculation.

\subsection{ Optimization Procedure} The objective function~\eqref{eq:objectivefn} is a convex function and can be efficiently optimized by  \emph{stochastic gradient descent} (SGD) procedure~\cite{shamir2013stochastic}. One computational issue associated with AUCL is that it pairs every positive and negative instance, effectively squaring the training set size.The SGD procedure easily overcomes this computational issue. At every step of SGD, a positive and a negative instance are randomly selected from the training set, followed by a gradient descent step. This makes the training procedure memory-efficient and mimics full gradient descent optimization on the entire loss. We also note that the rankers for the ads can be trained in parallel and any regularizer like $\|w\|_1$ and $\|w\|_2$ can be added to~\eqref{eq:objectivefn}, to introduce sparsity or avoid overfitting. Lastly, powerful non-linear kernel ranking functions can be learnt in place of linear ranking functions, but at the cost of memory efficiency, and the rankers can even be learnt online, from streaming data \cite{zhao2011online}.


\subsection{Constructing Policy from Rankers}
\label{rankerpolicy}

Similar to learning a classifier per ad, a separate ranking function $f_{w_a}(\cdot)$ is learnt for each ad $a$ from the bandit batch data. Then the following technique is used to convert the $K$ separate ranking functions into a recommendation policy. First, a threshold score $s_a$ is learnt for each action $a \in [K]$ separately (see the details below), and then for a new user $x$, the combined policy $\pi$ works as follows:
\begin{equation}
\label{eq:shifted-score}
\pi(x) =  \underset{a \in [K]}{\argmax}\ {(f_{w_a}(x)- s_a)}.
\end{equation}
Thus, $\pi$ maps $x$ to ad $a$ with maximum ``normalized score". This normalization negates the inherent scoring bias that might exist for each ranking function. That is, a ranking function for an action $a \in [K]$ might learn to score all instances (both positive and negative) higher than a ranking function for an action $b \in [K]$. Therefore, for a new instance $x$, ranking function for $a$ will always give a higher score than the ranking function for $b$, leading to possible incorrect predictions. 

{\bf Learning Threshold Score $s_a$}: After learning the ranking function $f_{w_a}(\cdot)$ from the training data, the threshold score $s_a$ is learnt by maximizing some classification measure like precision, recall, or F-score on the same training set. That is, score of each (positive or negative) instance in the training set is calculated and the classification measure corresponding to different thresholds are compared. The threshold that gives the maximum measure value is assigned to $s_a$. 
 

\section{Competing Policies and Evaluation Techniques}
To support our hypothesis that ranker based policies address the click-sparsity problem better than classifier based policies, we set up two sets of experiments. We a) compared deterministic policies (only ``exploitation") on full information (classification) datasets and b) compared stochastic policies (``exploitation + exploration") on bandit datasets, with a specific offline evaluation technique. Both of our experiments were designed for batch learning setting, with policies constructed from separate classifiers/rankers per ad. The classifiers considered were {\bf linear} and {\bf ensemble RandomForest} classifiers and ranker considered was the {\bf AUC optimized ranker}. 

{\bf Deterministic Policies}: Policies from the trained classifiers were constructed using the ``one-vs-all" technique, i.e.,~for a new user $x$, the ad with the maximum score according to the classifiers was predicted. For the policy constructed from rankers, the ad with the maximum shifted score according to the rankers was predicted, using Eq.~\ref{eq:shifted-score}. Deterministic policies are ``exploit only" policies.

{\bf Stochastic Policies}: Stochastic policies were constructed from deterministic policies by adding an $\epsilon$-greedy exploration technique on top. Briefly, let one of the stochastic policies be denoted by $\pi_e$ and let $\epsilon \in [0,1]$. For a context $x$ in the test set, $\pi_e(a | x)= 1- \epsilon$, if $a$ was the offer with the maximum score according to the underlying deterministic policy, and $\pi_e(a|x)= \frac{\epsilon}{K-1}$, otherwise ($K$ is the total number of offers). Thus, $\pi_e$ is a probability distribution over the offers. Stochastic policies are ``exploit+ explore" policies.

%
%

\subsection{Evaluation on Full Information Classification Data}
Benchmark bandit data are usually hard to obtain in public domains. So, we compared the deterministic policies on benchmark K-class classification data, converted to K-class bandit data, using the technique in~\cite{dudik2011doubly}. Briefly, the standard conversion technique is as follows: A $K$-class dataset is randomly split into training set $X_{\text{train}}$ and test set $X_{\text{test}}$ (in our experiments, we used $70-30$ split). \emph{Only the labeled training set is converted into bandit data, as per procedure}. Let $\{x,a\}$ be an instance and the corresponding class in the training set. A class $a' \in [K]$ is selected uniformly at random. If $a=a'$, a reward of $1$ is assigned to $x$; otherwise, a reward of $0$ is assigned. The new bandit instance is of the form $\{x,a',1\}$ or $\{x,a',0\}$, and the true class $a$ is hidden. The bandit data is then divided into $K$ separate binary class training sets, as detailed in the section ``Binary Classifier based Policies". \\
{\bf Evaluation Technique}: We compared the \emph{deterministic policies} by calculating the CTR of each policy. For a policy $\pi$, CTR on a test set of cardinality $n$ is measured as: 
\begin{equation}
\frac{1}{n} \sum_{(x,a) \in \text{test set}} \mathbbm{1}(\pi(x)=a)
\end{equation}
Note that we can calculate the true CTR of a policy $\pi$ because the correct class $a$ for an instance $x$ is known in the test set. 

\subsection{Evaluation on Bandit Information Data}
Bandit datasets have both training and test sets in bandit form, and datasets we use are industry proprietary in nature. \\
{\bf Evaluation Technique}: We compared the \emph{stochastic policies} on bandit datasets. Comparison of policies on bandit test set comes with the following unique challenge: for a stochastic policy $\pi$, the expected reward is $\rho(\pi)= \E_{a \sim \pi(\cdot|x)} r_{a}= \sum\limits_{a} r_{a} \pi(a|x)$, for a test context $x$ (with the true CTR of $\pi$ being average of expected reward over entire test set). \emph{Since the bandit form of test data does not give any information about rewards for offers which were not displayed, it is not possible to calculate the expected reward! } 

We evaluated the policies using a particular offline contextual bandit policy evaluation technique. There exist various such evaluation techniques in the literature, with adequate discussion about the process~\cite{li2011unbiased}. We used one of the importance weighted techniques as described in Theocharous et al. \shortcite{theocharous2015ad}. The reason was that we could give high confidence lower bound on the performance of the policies. We provide the mathematical details of the technique. 

The bandit test data was logged from the interaction between users and a fully random policy $\pi_u$, over an interaction window. The random policy produced the following distribution over offers: $\pi_u(a|x)= \frac{1}{K}$, $\forall \ a\in[K]$. For an instance $\{x,a',r_{a'}\}$ in the test set, the importance weighted reward of evaluation policy $\pi$ is computed as 
$\hat{\rho}(\pi)= r_{a'} \dfrac{\pi(a'|x)}{\pi_u(a'|x)}$.
The importance weighted reward is an unbiased estimator of the true expected reward of $\pi$, i.e.,~$\E_{a' \sim \pi_u(\cdot|x)} \hat{\rho}(\pi)= \rho(\pi)$. 

Let the cardinality of the test set be $n$. The {\bf importance weighted CTR of $\pi$} is defined as 
\begin{equation}
\dfrac{1}{n}\sum\limits_{i=1}^ n r_{a_i} \dfrac{\pi(a_i|x_i)}{\pi_u(a_i|x_i)}
\end{equation} 
Since $(x,r)$ are assumed to be generated i.i.d., the importance weighted CTR is an unbiased estimator of the true CTR of $\pi$. Moreover, it is possible to construct a \emph{t-test} based {\bf lower confidence bound} on the expected reward, using the unbiased estimator, as follows: let $X_i= r_{a_i} \dfrac{\pi(a_i|x_i)}{\pi_u(a_i|x_i)}$, $\hat{X}= \dfrac{1}{n} \sum\limits_{i=1}^n X_i$, and $\sigma= \sqrt{\dfrac{1}{n-1} \sum\limits_{i=1}^n (X_i - \hat{X})^2}$. Then, $\hat{X}= \text{importance weighted CTR}$ and 
\begin{equation}
\hat{X} - \dfrac{\sigma}{\sqrt{n}} t_{1-\delta, n-1}
\end{equation}
is a {\bf $1- \delta$ lower confidence bound} on the true CTR. Thus, during evaluation, we plotted the importance weighted CTR and lower confidence bounds for the competing policies.

\section{Empirical Results}
We detail the parameters and results of our experiments.\\
{\bf Linear Classifiers and Ranker}: For each ad $a$, a linear classifier was learnt by optimizing the logistic surrogate, while a linear ranker was learnt by optimizing the objective function~\eqref{eq:objectivefn}, with $\ell(\cdot)$ being the logistic surrogate. Since we did not have the problem of sparse high-dimensional features in our datasets, we added an $\ell_2$ regularizer instead of $\ell_1$ regularizer. We applied SGD with1 million iterations; varied the parameter $\lambda$ of the $\ell_2$ regularizer in the set $\{0.01,0.1,1,10\}$ and recorded the best result.\\
{\bf Ensemble Classifiers}: We learnt a RandomForest classifier for each ad $a$. The RandomForests were composed of 200 trees, both for computational feasibility and for the more theoretical reason outlined in~\cite{koh2014empirical}. 

\subsection{Comparison of Deterministic Policies}

{\bf Datasets}: The multi-class datasets are detailed in Table~\ref{tab:data}.
\begin{table*}[t]
\caption{All datasets were obtained from UCI repository (https://archive.ics.uci.edu/ml/). Five different datasets were selected. In the table, size represents the number of examples in the complete dataset. Features indicate the dimension of the instances. Avg. $\%$ positive gives the number of positive instances per class, divided by the total number of instances for that class, in the bandit training set, averaged over all classes. The lower the value, the more is the imbalance per class during training.} 
\label{tab:data}
\begin{center}
\begin{tabular}{cccccc}
\hline
& OptDigits & Isolet & Letter & PenDigits & Movementlibras \\
\hline
Size & 5620 & 7797 & 20000 & 10992& 360 \\
Features & 64& 617 & 16 & 16 & 91 \\
Classes & 10& 26 & 26 & 10& 15 \\
Avg. $\%$ positive & 10& 4 & 4 & 10 & 7\\
\hline
\end{tabular}
\end{center}
\end{table*}

\begin{figure}[!htb]
\begin{minipage}[b]{0.5\textwidth}
  \includegraphics[width=\textwidth]{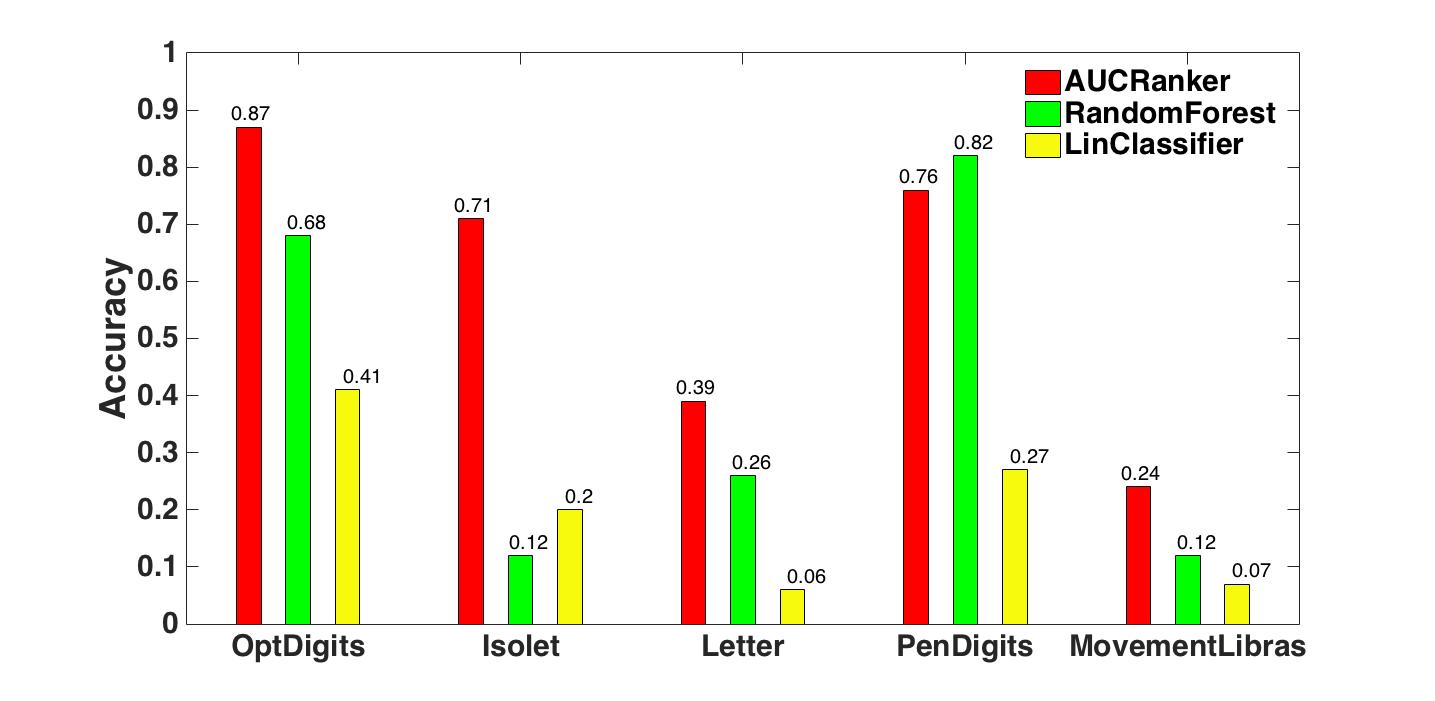}
\end{minipage}
\begin{minipage}[b]{0.5\textwidth}
  \includegraphics[width=\textwidth]{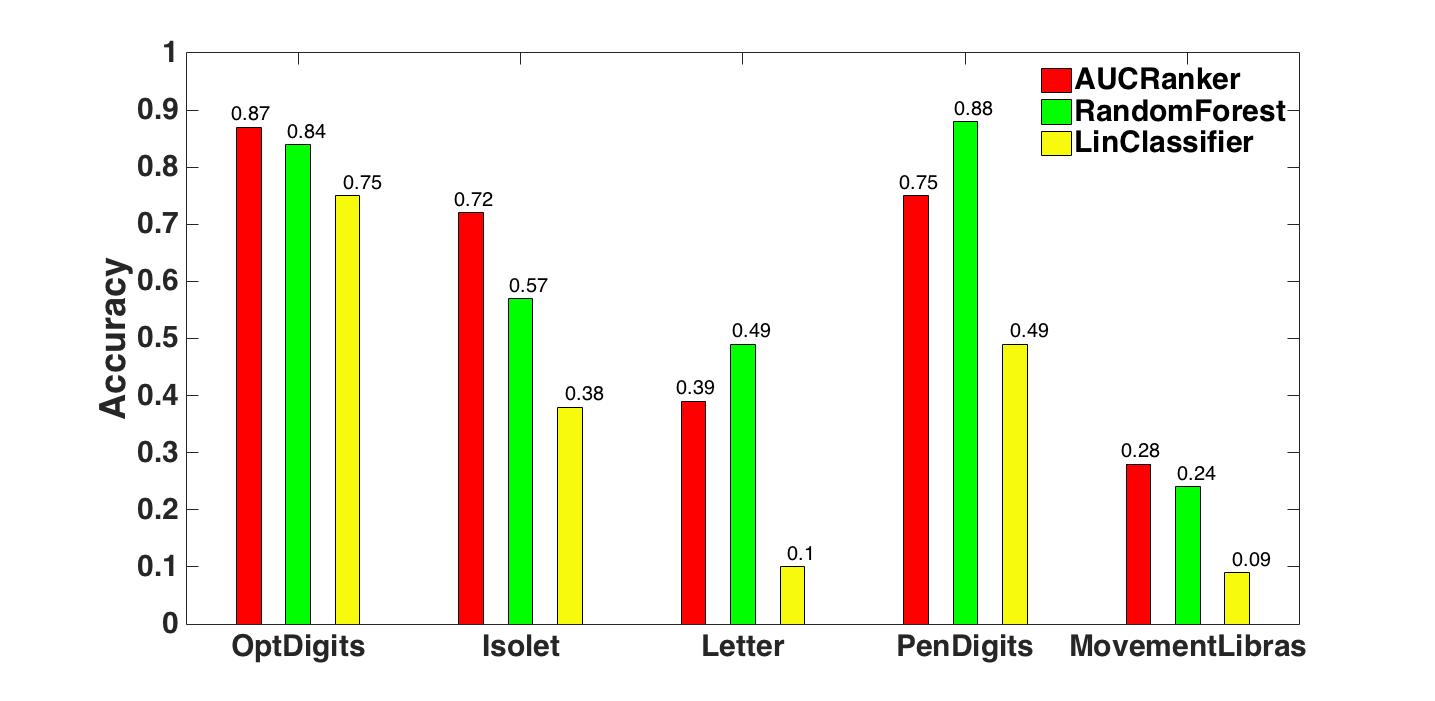}
\end{minipage}
\caption{Comparison of classification accuracy of various policies for different datasets, Top- {\bf without under-sampling} and Bottom- {\bf with under-sampling}. See {\bf Observations} for details}\label{Fig1}
\end{figure}

{\bf Evaluation}: To compare the deterministic policies , we conducted two sets of experiments; one without under-sampling of negative classes during training (i.e.,~no class balancing) and another with heavy under-sampling of negative classes (artificial class balancing). Training and testing were repeated 10 times for each dataset to account for the randomness introduced during conversion of classification training data to bandit training data, and the average accuracy over the runs are reported. Figure~\ref{Fig1} top and bottom show performance of various policies learnt without and with under-sampling during training, respectively. Under-sampling was done to make positive:negative ratio as 1:2 for every class (this basically means that Avg $\%$ positive was 33$\%$). \emph{The ratio of 1:2 generally gave the best results}. 

{\bf Observations}: {\bf a}) With heavy under-sampling, the performance of classifier-based policies improve significantly during training. Ranker-based policy is not affected, validating that class imbalance does not affect ranking loss, {\bf b)} The linear ranker-based policy performs uniformly better than the linear classifier-based policy, with or without under-sampling. This shows that restricting to same class of functions (linear), rankers handles class-imbalance much better than classifiers {\bf c)} The linear ranker-based policy does better than more complex RandomForest (RF) based policy, when no under-sampling is done during training, and is competitive when under-sampling is done, and {\bf d)} Complex classifiers like RFs are relatively robust to moderate class imbalance. However, as we will see in real datasets, when class imbalance is extreme, gain from using a ranker-based policy becomes prominent. Moreover, growing big forests may be infeasible due to memory constraints,

\begin{figure}[ht!]
     \begin{center}
     \subfigure[Hotel- {\bf moderate click sparsity} ]{%
            \label{Fig3}
           \includegraphics[height=35mm, width=90mm]{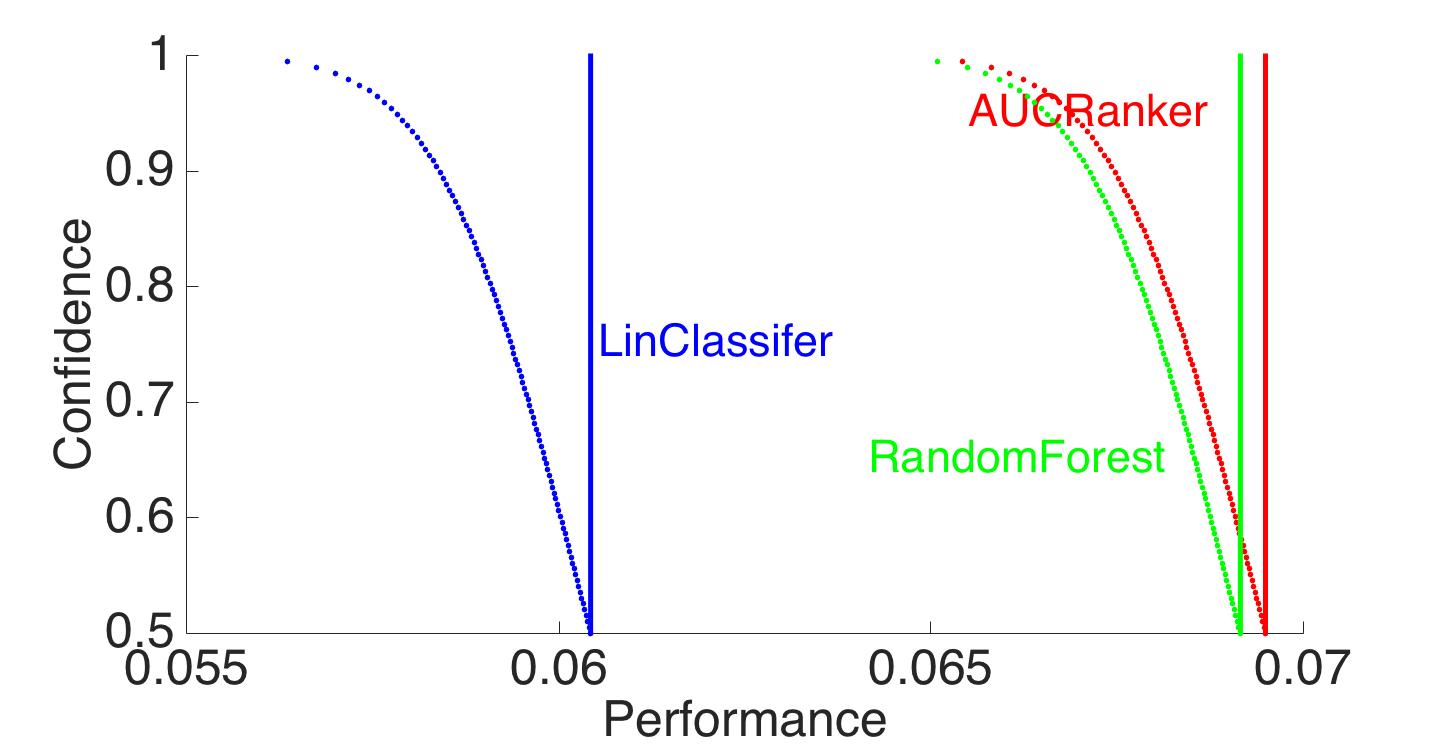}
        }\\
      \subfigure[Bank 1- {\bf extreme click sparsity}]{%
            \label{Fig4}
            \includegraphics[height=35mm, width=90mm]{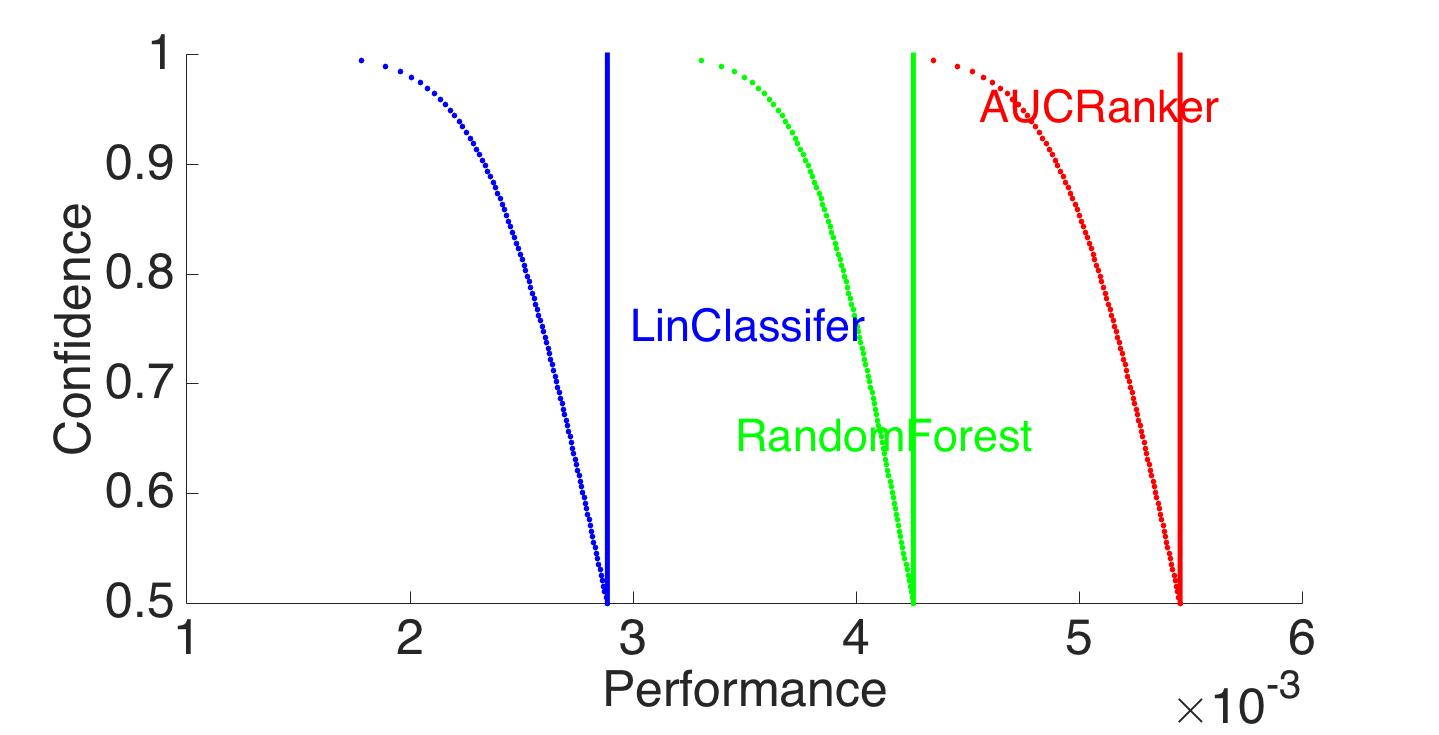}
        }\\
          \subfigure[Bank 2- {\bf extreme click sparsity}]{%
            \label{Fig5}
            \includegraphics[height=35mm, width=90mm]{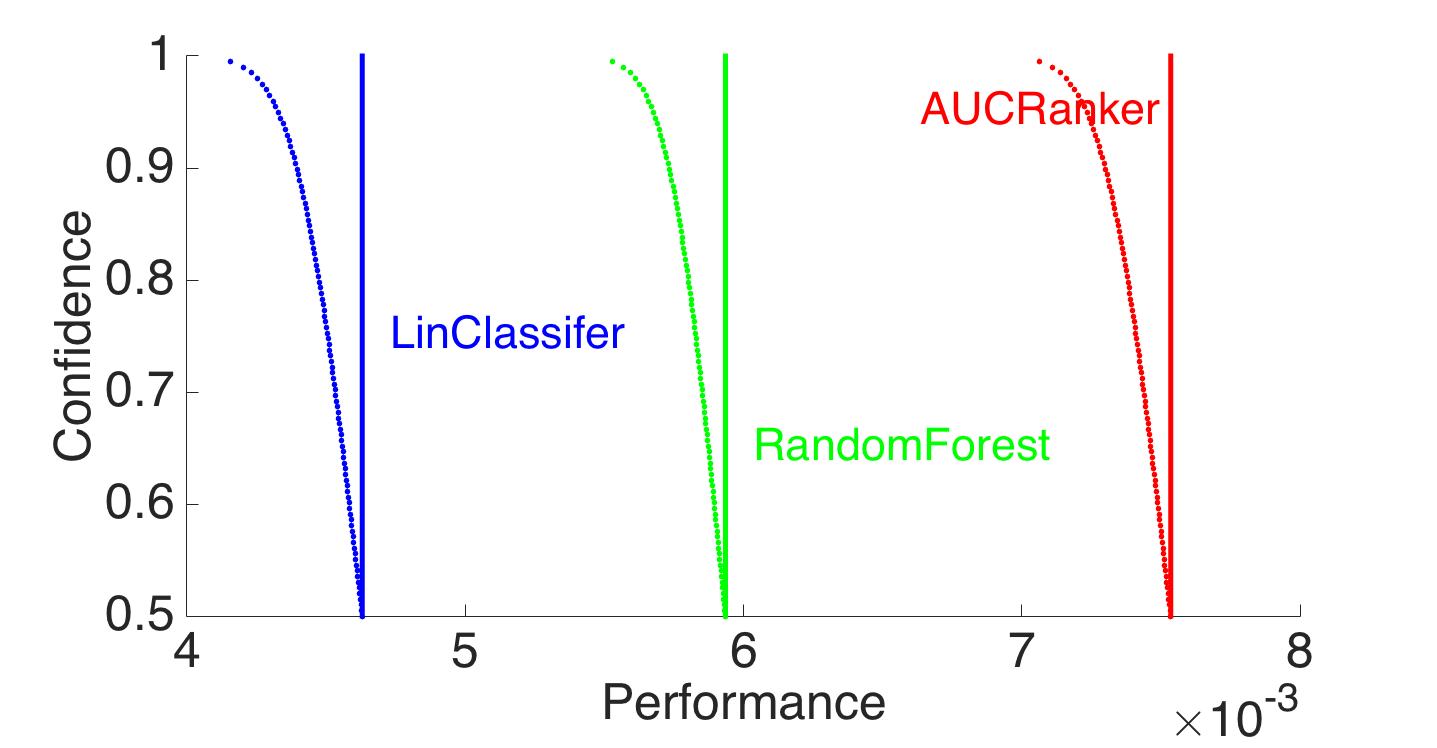}
        }%
      \end{center}
    \caption{Importance weighted CTR and lower confidence bounds for policies. Ranking based approach {\bf gains 15-25 $\%$ in importance weighted CTR} over RF based approach on data with {\bf extreme click sparsity}. Classifiers were trained after {\bf class balancing}.}%
   \label{fig:subfigures}
\end{figure}

\subsection{Comparison of Stochastic Policies}

Our next set of experiments were conducted on three different datasets that are property of a major technology company. 

{\bf Datasets}: Two of the datasets were collected from campaigns run by two major banks and another from campaign run by a major hotel chain. When a user visited the campaign website, she was either assigned to a targeted policy or a purely random policy. The targeted policy was some specific ad serving policy, particular to the campaign. The data was collected in the form $\{x,a,r_a\}$, where $x$ denotes the user context, $a$ denotes the offer displayed, and $r_a \in \{0,1\}$ denotes the reward received. We trained our competing policies on data collected from the targeted policy and testing was done on the data collected from the random policy. We focused on the top-5 offers by number of impressions in the training set. Table~\ref{tab:bandit} provides information about the training sets collected from the hotel and one of the bank's campaigns. The second bank's campaign has similar training set as the first one. \emph{As can be clearly observed, each offer in the bank's campaign suffers from extreme click sparsity}.

\begin{table}[t]
\caption{Information about the training sets}
\label{tab:bandit}
\begin{center}
\begin{tabular}{cccc}
\hline
Domain & Offer & Impressions & Avg. $\%$ Positive  \\
& & & (Clicks/Impressions)\\
\hline
Hotel & &&  \\
\hline
& 1 & 36164 & 8.1\\
& 2 & 37944 & 8.2 \\
& 3 & 30871 & 7.8\\
& 4 & 32765 & 7.7\\
& 5 & 20719 & 5.5\\
\hline
Bank 1 & &&\\
\hline
&1 & 37750& 0.17\\
&2 & 38254 & 0.40\\
&3 & 182191 & 0.45\\
&4 & 168789 & 0.30\\
&5 & 17291& 0.23\\
\hline
\end{tabular}
\end{center}
\end{table}

{\bf Feature Selection}: We used a feature selection strategy to select around 20$\%$ of the users' features, as some of the features were of poor quality and led to difficulty in learning. We used the \emph{information gain} criteria to select features~\cite{cheng2012fselector}.

{\bf Results}: Figures~\ref{Fig3},~\ref{Fig4}, and~\ref{Fig5} show the results of our experiments. We used heavy under-sampling of negative examples, at the ratio 1:1 for positive:negative examples per offer, while training the classifiers. During evaluation, $\epsilon=0.2$ was used as exploration parameter. Taking $\epsilon<0.2$, meaning more exploitation, did not yield better results. 

{\bf Observations}: {\bf a)} The ranker-based policy generally performed better than the classifier-based policies.
 {\bf b)} For the bank campaigns, \emph{where the click sparsity problem is extremely severe, it can be stated with high confidence that the ranker-based policy performed significantly better than classifier based policies}. This shows that the ranker-based policy can handle class imbalance better than the classifier policies.

\bibliography{References}

\begin{thebibliography}{}

\bibitem[\protect\citeauthoryear{Agarwal \bgroup et al\mbox.\egroup
  }{2009}]{agarwal2009translating}
Agarwal, D.; Gabrilovich, E.; Hall, R.; Josifovski, V.; and Khanna, R.
\newblock 2009.
\newblock Translating relevance scores to probabilities for contextual
  advertising.
\newblock In {\em Proceedings of the 18th ACM conference on Information and
  knowledge management},  1899--1902.
\newblock ACM.

\bibitem[\protect\citeauthoryear{Agarwal \bgroup et al\mbox.\egroup
  }{2014}]{agarwal2014taming}
Agarwal, A.; Hsu, D.; Kale, S.; Langford, J.; Li, L.; and Schapire, R.
\newblock 2014.
\newblock Taming the monster: A fast and simple algorithm for contextual
  bandits.
\newblock In {\em Proceedings of the 31st International Conference on Machine
  Learning},  1638--1646.

\bibitem[\protect\citeauthoryear{Agrawal and Goyal}{2013}]{agrawal2013thompson}
Agrawal, S., and Goyal, N.
\newblock 2013.
\newblock Thompson sampling for contextual bandits with linear payoffs.
\newblock In {\em Proceedings of the 30th International Conference on Machine
  Learning},  127--135.

\bibitem[\protect\citeauthoryear{Beygelzimer, Langford, and
  Ravikumar}{2007}]{beygelzimer2007multiclass}
Beygelzimer, A.; Langford, J.; and Ravikumar, P.
\newblock 2007.
\newblock Multiclass classification with filter trees.
\newblock {\em Preprint, June} 2.

\bibitem[\protect\citeauthoryear{Calders and
  Jaroszewicz}{2007}]{calders2007efficient}
Calders, T., and Jaroszewicz, S.
\newblock 2007.
\newblock Efficient {AUC} optimization for classification.
\newblock In {\em Knowledge Discovery in Databases: PKDD 2007}. Springer.
\newblock  42--53.

\bibitem[\protect\citeauthoryear{Cao, Zhao, and
  Zaiane}{2013}]{cao2013optimized}
Cao, P.; Zhao, D.; and Zaiane, O.
\newblock 2013.
\newblock An optimized cost-sensitive svm for imbalanced data learning.
\newblock In {\em Advances in Knowledge Discovery and Data Mining}. Springer.
\newblock  280--292.

\bibitem[\protect\citeauthoryear{Chapelle and Li}{2011}]{chapelle2011empirical}
Chapelle, O., and Li, L.
\newblock 2011.
\newblock An empirical evaluation of thompson sampling.
\newblock In {\em Advances in neural information processing systems},
  2249--2257.

\bibitem[\protect\citeauthoryear{Chawla, Japkowicz, and
  Kotcz}{2004}]{chawla2004editorial}
Chawla, N.; Japkowicz, N.; and Kotcz, A.
\newblock 2004.
\newblock Editorial: special issue on learning from imbalanced data sets.
\newblock {\em ACM Sigkdd Explorations Newsletter} 6(1):1--6.

\bibitem[\protect\citeauthoryear{Cheng, Wang, and
  Bryant}{2012}]{cheng2012fselector}
Cheng, T.; Wang, Y.; and Bryant, S.
\newblock 2012.
\newblock {FS}elector: a ruby gem for feature selection.
\newblock {\em Bioinformatics} 28(21):2851--2852.

\bibitem[\protect\citeauthoryear{Chu \bgroup et al\mbox.\egroup
  }{2011}]{chu2011contextual}
Chu, W.; Li, L.; Reyzin, L.; and Schapire, R.
\newblock 2011.
\newblock Contextual bandits with linear payoff functions.
\newblock In {\em International Conference on Artificial Intelligence and
  Statistics},  208--214.

\bibitem[\protect\citeauthoryear{Cortes and Mohri}{2004}]{cortes2004auc}
Cortes, C., and Mohri, M.
\newblock 2004.
\newblock {AUC} optimization vs. error rate minimization.
\newblock {\em Advances in neural information processing systems}
  16(16):313--320.

\bibitem[\protect\citeauthoryear{Dudik \bgroup et al\mbox.\egroup
  }{2011}]{dudik2011efficient}
Dudik, M.; Hsu, D.; Kale, S.; Karampatziakis, N.; Langford, J.; Reyzin, L.; and
  Zhang, T.
\newblock 2011.
\newblock Efficient optimal learning for contextual bandits.
\newblock {\em Proceedings of the 27th Conference on Uncertainty in Artificial
  Intelligence, 2011}.

\bibitem[\protect\citeauthoryear{He and others}{2014}]{he2014practical}
He, X., et~al.
\newblock 2014.
\newblock Practical lessons from predicting clicks on ads at facebook.
\newblock In {\em Proceedings of 20th ACM SIGKDD Conference on Knowledge
  Discovery and Data Mining},  1--9.
\newblock ACM.

\bibitem[\protect\citeauthoryear{Japkowicz and
  Stephen}{2002}]{japkowicz2002class}
Japkowicz, N., and Stephen, S.
\newblock 2002.
\newblock The class imbalance problem: A systematic study.
\newblock {\em Intelligent data analysis} 6(5):429--449.

\bibitem[\protect\citeauthoryear{Koh and Gupta}{2014}]{koh2014empirical}
Koh, E., and Gupta, N.
\newblock 2014.
\newblock An empirical evaluation of ensemble decision trees to improve
  personalization on advertisement.
\newblock In {\em Proceedings of KDD 14 Second Workshop on User Engagement
  Optimization}.

\bibitem[\protect\citeauthoryear{Langford and Zhang}{2008}]{langford2008epoch}
Langford, J., and Zhang, T.
\newblock 2008.
\newblock The epoch-greedy algorithm for multi-armed bandits with side
  information.
\newblock In {\em Advances in neural information processing systems},
  817--824.

\bibitem[\protect\citeauthoryear{Langford, Li, and
  Dudik}{2011}]{dudik2011doubly}
Langford, J.; Li, L.; and Dudik, M.
\newblock 2011.
\newblock Doubly robust policy evaluation and learning.
\newblock In {\em Proceedings of the 28th International Conference on Machine
  Learning},  1097--1104.

\bibitem[\protect\citeauthoryear{Li \bgroup et al\mbox.\egroup
  }{2011}]{li2011unbiased}
Li, L.; Chu, W.; Langford, J.; and Wang, X.
\newblock 2011.
\newblock Unbiased offline evaluation of contextual-bandit-based news article
  recommendation algorithms.
\newblock In {\em Proceedings of the fourth ACM international conference on Web
  search and data mining},  297--306.
\newblock ACM.

\bibitem[\protect\citeauthoryear{McMahan and others}{2013}]{mcmahan2013ad}
McMahan, H., et~al.
\newblock 2013.
\newblock Ad click prediction: a view from the trenches.
\newblock In {\em Proceedings of the 19th ACM SIGKDD international conference
  on Knowledge discovery and data mining},  1222--1230.
\newblock ACM.

\bibitem[\protect\citeauthoryear{Richardson, Dominowska, and
  Ragno}{2007}]{richardson2007predicting}
Richardson, M.; Dominowska, E.; and Ragno, R.
\newblock 2007.
\newblock Predicting clicks: estimating the click-through rate for new ads.
\newblock In {\em Proceedings of the 16th international conference on World
  Wide Web},  521--530.
\newblock ACM.

\bibitem[\protect\citeauthoryear{Rifkin and Klautau}{2004}]{rifkin2004defense}
Rifkin, R., and Klautau, A.
\newblock 2004.
\newblock In defense of one-vs-all classification.
\newblock {\em The Journal of Machine Learning Research} 5:101--141.

\bibitem[\protect\citeauthoryear{Shamir and Zhang}{2013}]{shamir2013stochastic}
Shamir, O., and Zhang, T.
\newblock 2013.
\newblock Stochastic gradient descent for non-smooth optimization: Convergence
  results and optimal averaging schemes.
\newblock In {\em Proceedings of the 30th International Conference on Machine
  Learning, 2013},  71--79.

\bibitem[\protect\citeauthoryear{Theocharous, Thomas, and
  Ghavamzadeh}{2015}]{theocharous2015ad}
Theocharous, G.; Thomas, P.; and Ghavamzadeh, M.
\newblock 2015.
\newblock Ad recommendation systems for life-time value optimization.
\newblock In {\em Proceedings of the 24th International Conference on World
  Wide Web Companion},  1305--1310.

\bibitem[\protect\citeauthoryear{Thomas, Theocharous, and
  Ghavamzadeh}{2015}]{Thomas15HC}
Thomas, P.; Theocharous, G.; and Ghavamzadeh, M.
\newblock 2015.
\newblock High confidence off-policy evaluation.
\newblock In {\em Proceedings of the Twenty-Ninth Conference on Artificial
  Intelligence}.

\bibitem[\protect\citeauthoryear{Zhao \bgroup et al\mbox.\egroup
  }{2011}]{zhao2011online}
Zhao, P.; Jin, R.; Yang, T.; and Hoi, S.~C.
\newblock 2011.
\newblock Online auc maximization.
\newblock In {\em Proceedings of the 28th International Conference on Machine
  Learning (ICML-11)},  233--240.

\end{thebibliography}
\bibliographystyle{ijcai16}

\end{document}